\documentclass[10pt,twocolumn,letterpaper]{article}
\usepackage{abstract}
\usepackage[pagebackref=true,breaklinks=true,letterpaper=true,colorlinks,bookmarks=false]{hyperref}
\usepackage{cvpr}
\usepackage{times}
\usepackage{epsfig}
\usepackage{graphicx}
\usepackage{amsmath}
\usepackage{amssymb}
\usepackage{bbm}
\usepackage{comment}
\usepackage[ruled,lined]{algorithm2e}
\usepackage{makecell}
\usepackage{multirow}
\usepackage{subcaption}
\usepackage{float}
\usepackage{amsmath,amsfonts,bm}
\usepackage{cleveref}

\newcommand{\real}{\mathbb{R}}

\newcommand{\context}{\mathcal{H}}

\DeclareMathOperator*{\argmax}{argmax}

\DeclareMathAlphabet{\mathsfit}{\encodingdefault}{\sfdefault}{m}{sl}
\SetMathAlphabet{\mathsfit}{bold}{\encodingdefault}{\sfdefault}{bx}{n}

\newcommand{\eat}[1]{} 

\newcommand{\fancyname}{Multiverse}

\cvprfinalcopy 

\def\cvprPaperID{2474} 

\ifcvprfinal\fi
\begin{document}

\title{
The Garden of Forking Paths:
Towards Multi-Future Trajectory Prediction 
}

\author{
Junwei Liang\textsuperscript{1}\thanks{Work partially done during a research internship at Google.} \qquad
Lu Jiang\textsuperscript{2} \qquad
Kevin Murphy\textsuperscript{2} \qquad
Ting Yu\textsuperscript{3} \qquad
Alexander Hauptmann\textsuperscript{1} \\
\textsuperscript{1}Carnegie Mellon University \qquad\qquad \textsuperscript{2}Google Research \qquad\qquad
\textsuperscript{3}Google Cloud AI\\
{\tt\small \{junweil,alex\}@cs.cmu.edu, \{lujiang,kpmurphy,yuti\}@google.com} 
\vspace{-.5em}\\
}

\twocolumn[{%
\renewcommand\twocolumn[1][]{#1}%
\maketitle
\begin{center}
\centering
\vspace{-10mm}
   \includegraphics[width=1.0\textwidth]{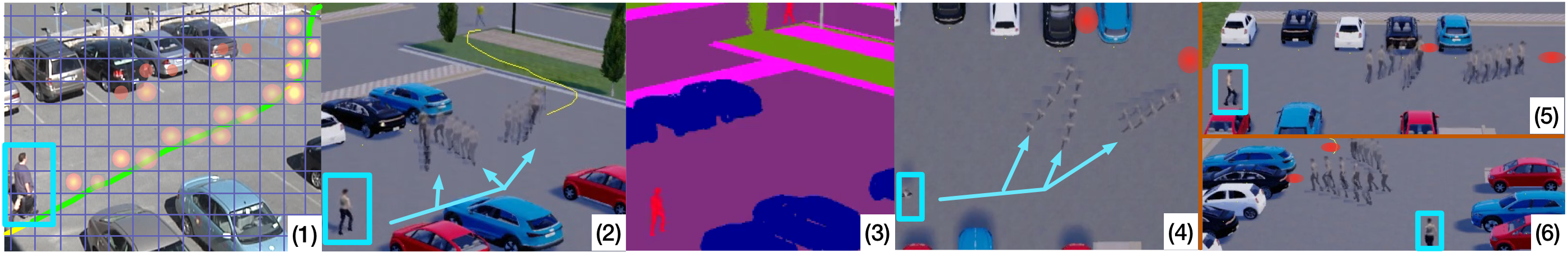}
   \vspace{-7mm}
    \captionof{figure}{
    Illustration of person trajectory prediction.
    (1) A person walks towards a car (data from
    the VIRAT/ActEV dataset).
    The green line is the actual future trajectory and the yellow-orange heatmaps are example future predictions.
    Although these predictions near the cars
    are plausible, they would be considered errors in
    the real video dataset. 
    (2) To combat this,
     we propose a new dataset called ``Forking Paths'';
     here we illustrate 3 possible futures
     created by human annotators controlling agents
      in a synthetic world derived from real data.
     (3) Here we show semantic segmentation of the scene.
     (4-6) Here we show the same scene rendered from different viewing angles, where the red circles are future destinations.
    \eat{
    Our goal is to predict a
    distribution
    over a person's future trajectory. The first from the left image shows a person walking in a parking lot from a real video, where the green line is the actual future trajectory and the yellow-orange heatmaps are example future predictions. As we see, the model predictions around the cars are plausible but will be considered errors in real video dataset. Therefore we propose a new dataset called ``Forking Paths'' dataset that is reconstructed from real-world video and with human-annotated possible future paths. We show our dataset from different viewing angles and with overlaid future frames of the person. The middle image shows pixel-accurate scene semantic segmentation ground truth of our dataset.
    }
    \vspace{-1mm}
    }
    \label{fig:multiFutures}
\end{center}%
}]
\saythanks


\setlength{\abovedisplayskip}{2pt} \setlength{\belowdisplayskip}{3pt}

\vspace{-2mm}
\begin{abstract}
\vspace{-2mm}
This paper studies the problem of predicting the distribution over multiple possible future paths of people as they move through various visual scenes.
We make two main contributions.
The first contribution is a new dataset, created in a realistic 3D simulator,
which is based on real world trajectory data, and then
extrapolated by human annotators to achieve different latent goals.
This provides the first benchmark for quantitative evaluation
of the models to predict multi-future trajectories.
The second contribution is a new model to generate multiple plausible future trajectories,
which contains novel designs of using multi-scale location encodings and convolutional RNNs over graphs.
We refer to our model as \textit{\fancyname}. 
We show that our model achieves the best results
on our dataset, as well as on the real-world VIRAT/ActEV dataset
(which just contains one possible future).
~\footnote{Code and models are released at \url{https://next.cs.cmu.edu/multiverse}}
\end{abstract}
\vspace{-7mm}

\vspace{-1mm}
\section{Introduction}
\vspace{-1mm}
Forecasting future
human behavior is a fundamental problem in video understanding. In particular, future path prediction, which aims at forecasting a pedestrian's future trajectory in the next few seconds, has received 
a lot of attention in our community~\cite{kitani2012activity,alahi2016social,gupta2018social,li2019way}. This functionality 
is a key component in a variety of applications such as autonomous driving~\cite{bansal2018chauffeurnet,chai2019multipath}, long-term object tracking~\cite{kalman1960new, sadeghian2017tracking}, safety monitoring~\cite{liang2019peeking}, robotic planning~\cite{rhinehart2017first,rhinehart2018r2p2},
\etc.

Of course, the future is often very uncertain:
Given the same historical trajectory, a person
may take different paths, depending on their 
(latent) goals. Thus recent work has started focusing on  \emph{multi-future trajectory prediction}~\cite{tang2019multiple,chai2019multipath,li2019way,makansi2019overcoming, thiede2019analyzing,lee2017desire}.

Consider the example in Fig.~\ref{fig:multiFutures}.
We see a person moving from the bottom left towards the top right of
the image, and our task is to predict where he will go next.
Since there are many possible future trajectories this person
might follow, we are interested in learning a model
that can generate multiple plausible futures.
However, since the ground truth data only contains one trajectory, it is difficult to evaluate such
probabilistic models.
\eat{
However as we see in Fig.~\ref{fig:multiFutures}, where the green line is the ground truth future trajectory and the orange heatmaps are example future predictions, the plausible predictions around the cars will not be considered correct in existing datasets.
}

To overcome the aforementioned challenges,
our first contribution is the creation of a realistic synthetic dataset that allows us to compare models in a quantitative way in terms of their ability to predict multiple plausible futures,
rather than just evaluating them against a single observed trajectory as in existing studies.
We create this dataset using the
3D CARLA~\cite{dosovitskiy2017carla} simulator, where the scenes are manually designed to be similar to those found in the  challenging real-world benchmark VIRAT/ActEV~\cite{oh2011large,2018trecvidawad}.
Once we have recreated the static scene, we automatically reconstruct trajectories by projecting real-world data to the 3D simulation world. See Fig.~\ref{fig:multiFutures} and~\ref{fig:dataset}.
We then semi-automatically select a set of plausible future destinations (corresponding to semantically meaningful locations in the scene), and ask human annotators to create multiple possible
continuations of the real trajectories towards each such goal. In this way, our dataset is ``anchored'' in reality,
and yet contains plausible variations in high-level human behavior, which is impossible to simulate automatically.

We call this dataset the ``Forking Paths'' dataset,
a reference to the short story by Jorge Luis Borges.\footnote{
\footnotesize{\url{https://en.wikipedia.org/wiki/The_Garden_of_Forking_Paths}}
} %
As shown in Fig.~\ref{fig:multiFutures}, different human annotations have created forkings of future trajectories for the identical historical past.
So far, we have collected 750 sequences, with each covering about 15 seconds, 
from 10 annotators, controlling 127 agents in 7 different scenes. 
Each agent contains 5.9 future trajectories on average.
We render each sequence from 4 different views,
and automatically generate dense labels,
as illustrated in Fig.~\ref{fig:multiFutures} and \ref{fig:dataset}.
In total,
this amounts to 3.2 hours of trajectory sequences, which is comparable to the largest person trajectory benchmark VIRAT/ActEV~\cite{2018trecvidawad, oh2011large} (4.5 hours), or 5 times bigger than the common ETH/UCY~\cite{lerner2007crowds,luber2010people} benchmark.
We therefore believe this will serve as a benchmark
for evaluating models that can predict multiple futures.

Our second contribution is to propose a new probabilistic model, \textit{\fancyname}, which can generate multiple plausible trajectories given the past history of locations and the scene.
The model contains two novel design decisions. First, we use a multi-scale representation of locations. 
In the first scale, the coarse scale, we represent locations on a 2D grid, as shown in Fig.~\ref{fig:multiFutures}(1). This captures high level uncertainty about possible destinations and leads to a better representation of multi-modal distributions. In the second fine scale, we predict a real-valued offset for each grid cell, to get more precise localization. This two-stage approach is partially inspired by object detection methods~\cite{ren2015faster}. 
The second novelty of our model is to design convolutional RNNs~\cite{xingjian2015convolutional} over the spatial graph as a way of encoding inductive bias about the movement patterns of people.

In addition, we empirically validate our model on the challenging real-world benchmark VIRAT/ActEV~\cite{oh2011large,2018trecvidawad} for single-future trajectory prediction, in which our model achieves the best-published result. On the proposed simulation data for multi-future prediction, experimental results show our model compares favorably against the state-of-the-art models across different settings. To summarize, the main contributions of this paper are as follows:
\textit{(i)} We introduce the first dataset and evaluation methodology that allows us to compare models in a quantitative way in terms of their ability to predict multiple plausible futures.
\textit{(ii)} We propose a new effective model for multi-future trajectory prediction.
\textit{(iii)} We establish a new
state of the art result
on the challenging VIRAT/ActEV benchmark,
and compare
various methods
on our multi-future prediction dataset. 
\vspace{-4mm}

\section{Related Work}
\vspace{-1mm}

\noindent\textbf{Single-future trajectory prediction.}
Recent works have tried to predict a single best trajectory for pedestrians or vehicles.
Early works~\cite{manh2018scene, xue2018ss,zhang2019sr} focused on modeling person motions by considering them as points in the scene.
These research works~\cite{kooij2014context,yagi2018future,ma2017forecasting,liang2019peeking} have attempted to predict person paths by utilizing visual features. 
Recently Liang \etal ~\cite{liang2019peeking} proposed a joint future activity and trajectory prediction framework that utilized multiple visual features using focal attention~\cite{liang2018focal,liang2019focal}.
Many works~\cite{lee2017desire,sadeghian2018car,bansal2018chauffeurnet,hong2019rules,zhao2019multi} in vehicle trajectory prediction have been proposed.
CAR-Net~\cite{sadeghian2018car} proposed attention networks on top of scene semantic CNN to predict vehicle trajectories.
Chauffeurnet~\cite{bansal2018chauffeurnet} utilized imitation learning for trajectory prediction.

\noindent\textbf{Multi-future trajectory prediction.}
Many works have tried to model the uncertainty of trajectory prediction.
Various papers (\eg \cite{kitani2012activity,rhinehart2018r2p2,rhinehart2019precog} use Inverse Reinforcement Learning (IRL) to forecast human trajectories. 
Social-LSTM~\cite{alahi2016social} is a popular method using social pooling to predict future trajectories.
Other works~\cite{sadeghian2018sophie,gupta2018social,li2019way,amirian2019social} like Social-GAN~\cite{gupta2018social} have utilized generative adversarial networks~\cite{goodfellow2014generative} to generate diverse person trajectories.
In vehicle trajectory prediction, DESIRE~\cite{lee2017desire} utilized variational auto-encoders (VAE) to predict future vehicle trajectories.
Many recent works~\cite{thiede2019analyzing,chai2019multipath,tang2019multiple,makansi2019overcoming} also proposed probabilistic frameworks for multi-future vehicle trajectory prediction. 
Different from these previous works, we present a flexible two-stage framework that combines multi-modal distribution modeling and precise location prediction.

\noindent\textbf{Trajectory Datasets.}
Many vehicle trajectory datasets~\cite{caesar2019nuscenes,chang2019argoverse} have been proposed as a result of self-driving's surging popularity. 
With the recent advancement in 3D computer vision research~\cite{zhang2015fast,liang2017event,shah2018airsim,dosovitskiy2017carla,richter2016playing,ros2016synthia,heess2017emergence}, many research works~\cite{qiu2017unrealcv, gaidon2016virtual,de2017procedural,das2018embodied,wu2019revisiting,zhu2017target,sun2019stochastic} have looked into 3D simulated environment for its flexibility and ability to generate enormous amount of data.
We are the first to propose a 3D simulation dataset that is reconstructed from real-world scenarios complemented with a variety of human trajectory continuations for multi-future person trajectory prediction. 

\begin{figure*}[ht]
	\centering
	    \vspace{-4mm}
		\includegraphics[width=\textwidth]{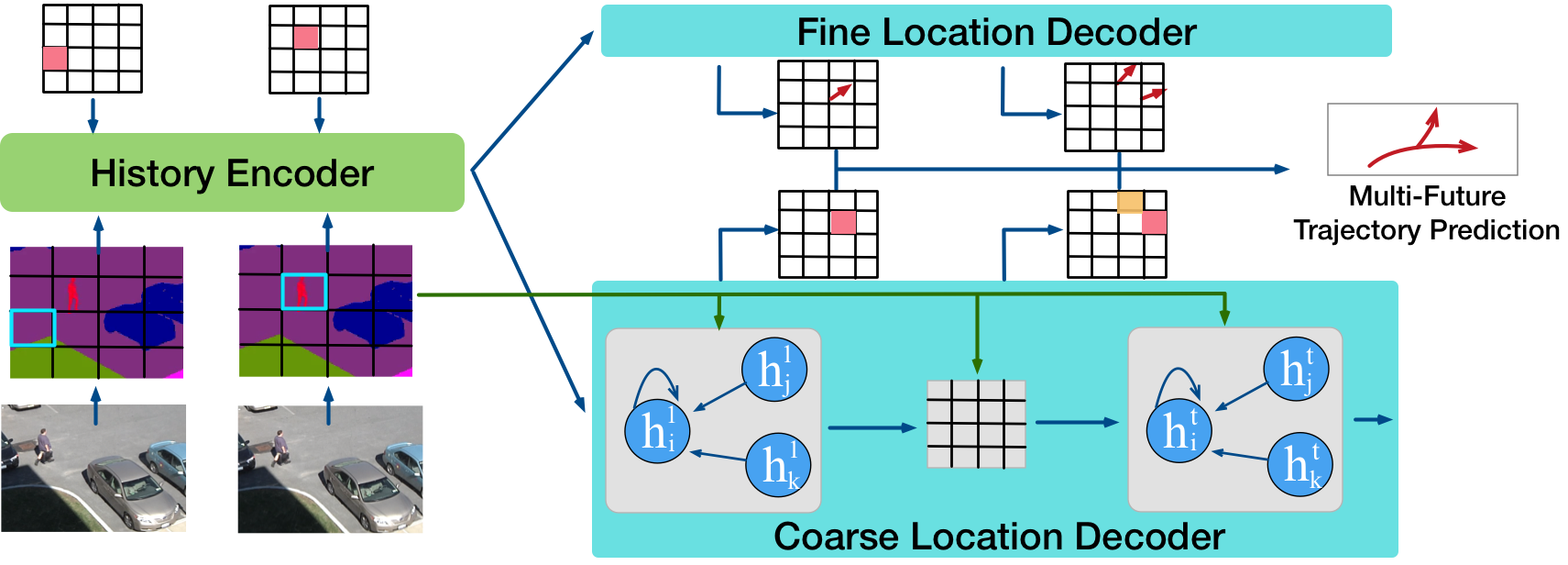}
		\vspace{-8mm}
	\caption{Overview of our model.
	The input to the model is 
	the ground truth location history,
	and a set of video frames,
	which are preprocessed by a semantic segmentation model. 
	This is encoded by the ``History Encoder''
	convolutional RNN.
	The output of the encoder is fed
	to the convolutional RNN decoder for  location prediction.
	The coarse location decoder outputs
	a heatmap over the 2D grid of size $H \times W$.
	The fine location decoder
	outputs 
	a vector offset within each grid cell.
	These are combined to generate a multimodal distribution
	over $\real^2$ for predicted locations.
	}
	\label{fig:model}
	\vspace{-5mm}
\end{figure*}

\vspace{-3mm}
\section{Methods}
\vspace{-2mm}
\label{sec:method}
\label{sec:approach}

In this section, we describe our 
model for forecasting agent
trajectories,
which we call \textit{\fancyname}.
We focus on predicting the locations of a single agent for multiple steps into the future, $L_{h+1:T}$, 
given a sequence of past video frames, $V_{1:h}$,
and agent locations, $L_{1:h}$,
where $h$ is the history length and $T-h$ is the prediction length.
Since there is inherent uncertainty in this task, our goal is to design a model
that can effectively predict multiple plausible future trajectories,
by computing the multimodal distribution
$p(L_{h+1:T}|L_{1:h}, V_{1:h})$.
See Fig.~\ref{fig:model} for a high level summary of the model,
and the sections below for more details.

\vspace{-1mm}
\subsection{History Encoder}
\vspace{-1mm}
The encoder computes a representation of the scene
from the history of past locations,
$L_{1:h}$, and frames, $V_{1:h}$.
We encode each ground truth location
$L_t$ by an index $Y_t \in G$ representing
the nearest cell in a 2D grid $G$ of size $H \times W$, indexed from $1$ to $HW$.
 Inspired by~\cite{lazebnik2006beyond,lin2017feature}, 
 we encode location with two different grid scales
 ($36 \times 18$ and $18 \times 9$);
 we show the benefits of this multi-scale
 encoding in Section~\ref{sec:ablation}.
For simplicity of presentation,
we focus on a  single $H \times W$ grid.

To make the model more invariant to low-level visual details, and thus more robust to domain shift (\eg, between different scenes, different views of the same scene, or between real and synthetic images), we preprocess each video frame $V_t$ using a pre-trained semantic segmentation model, with $K=13$ possible class labels per pixel. 
We use the Deeplab model \cite{chen2017deeplab} trained on the ADE20k~\cite{zhou2017scene} dataset,
and keep its weights frozen.
Let $S_t$ be this semantic segmentation map modeled as a tensor of size $H \times W \times K$.

We then pass these inputs to a convolutional RNN 
\cite{xingjian2015convolutional,wang2019eidetic}
to compute a spatial-temporal feature history:
\begin{equation}
H_t^e = \text{ConvRNN}(\text{one-hot}(Y_{t}) \odot
(W * S_t), H^e_{t-1})
\end{equation}
where
$\odot$ is element wise product, 
and $*$ represents 2D-convolution. The function $\text{one-hot}(\cdot)$ projects a cell index into an one-hot embedding of size $H \times W$ according to its spatial location.
\eat{ 
We concatenate the final state of this encoder,
together with the temporal average of the semantic maps,
$\overline{S} = \frac{1}{h} \sum_{t=1}^h S_t$,
as the representation of the context $\context=[H_h^e,\overline{S}]$.
}
We use the final state of this encoder $H_t^e \in \mathbb{R}^{H \times W \times d_{enc}}$, where $d_{enc}$ is the hidden size, to initialize the state of the decoders. 
We also use the temporal average of the semantic maps,
$\overline{S} = \frac{1}{h} \sum_{t=1}^h S_t$, during each decoding step.
The context is represented as $\context=[H_h^e,\overline{S}]$.

\vspace{-1mm}
\subsection{Coarse Location Decoder}
\vspace{-1mm}
After getting the context $\context$, our goal is to forecast future locations.
We initially focus on predicting locations
at the level of grid cells, $Y_t \in G$.
In Section~\ref{sec:fineDecoder}, 
we discuss how to predict a continuous offset in $\real^2$, which specifies a ``delta''
from the center of each grid cell, to get a fine-grained location prediction.

\eat{
\begin{align}
\begin{split}
    p(Y_t=j|\context) &= \sum_{i \in G}
    p(Y_{t}=j|Y_{t-1}=i,\context) p(Y_{t-1}=i|\context) 
\end{split}
\end{align}
}
Let the coarse
distribution over grid
locations at time $t$ (known as the ``belief state'') be denoted by
$C_t(i)=p(Y_t=i|Y_{h:t-1},\context)$,
for $\forall i \in G$ and $t \in [h+1, T]$. For brevity, we use a single
index $i$ to represent a cell in the 2D grid.
Rather than assuming a Markov model,
we update this using a 
convolutional recurrent neural network,
with hidden states $H_t^C$.
We then compute the belief state by:
\begin{align}
    C_t = \text{softmax}(W * H_t^C) \in \mathbb{R}^{HW}
    \label{eqn:fC}
\end{align}
Here we use 2D-convolution with one filter and flatten the spatial dimension before applying softmax.
The hidden state is updated using:
\begin{align}
    H_t^C = \text{ConvRNN}(\text{GAT}(H^C_{t-1}), 
    \text{embed}(C_{t-1}))
    \label{eqn:fH}
\end{align}
where
$\text{embed}(C_{t-1})$ embeds into a 3D tensor of size $H \times W \times d_e$ and $d_e$ is the embedding size.
$\text{GAT}(H_{t-1}^C)$ is a 
graph attention network~\cite{velivckovic2017graph},
where the graph structure corresponds to the 2D grid in $G$.
More precisely,
let $h_i$ be the feature vector corresponding
to the $i$-th grid cell in $H^C_{t-1}$,
and let $\tilde{h}_i$ be the 
corresponding output
in
$\tilde{H}^C_{t-1} = \text{GAT}(H^C_{t-1}) \in \mathbb{R}^{H \times W \times d_{dec}}$, where $d_{dec}$ is the size of the decoder hidden state.
We compute these outputs of GAT using:
\begin{align}
    \tilde{h}_i = \frac{1}{|\mathcal{N}_i|}
    \sum_{j \in \mathcal{N}_i} f_e([v_i, v_j]) + h_i
\end{align}
where $\mathcal{N}_i$ are the neighbors of node $v_i$ in $G$ with each node represented as 
$v_i = [h_i, \overline{S}_{i}]$, where $\overline{S}_{i}$ collects the cell $i$'s feature in $\overline{S}$.
$f_e$ is some edge function (implemented as an MLP in our experiments) that computes the attention weights.

The graph-structured
update function for the RNN ensures that the probability
mass  ``diffuses out'' to nearby grid cells
in a controlled manner, reflecting the prior knowledge
that people do not suddenly jump between distant locations.
This inductive bias is also encoded in the convolutional structure,
but adding the graph attention network gives improved results,
because the weights are input-dependent and not fixed.

\vspace{-1mm}
\subsection{Fine Location Decoder}
\label{sec:fineDecoder}
\vspace{-1mm}

The 2D heatmap is useful for capturing multimodal distributions, but does not give very precise location predictions. 
To overcome this, we train a second convolutional RNN decoder
$H_t^O$
to compute an offset vector for each possible
grid cell
using a regression output, $O_t = \text{MLP}(H_t^O) \in \mathbb{R}^{H \times W \times 2}$.
This RNN is updated using
\begin{align}
    H_t^O = \text{ConvRNN}(\text{GAT}(H^O_{t-1}), O_{t-1}) \in \mathbb{R}^{H \times W \times d_{dec}}
\end{align}

\noindent To compute the final prediction location, we first flatten the spatial dimension of $O_{t}$ into $\tilde{O}_{t} \in \mathbb{R}^{HW \times 2}$. Then we use
\begin{align}
    L_t = Q_{i} + \tilde{O}_{ti}
\label{eq:L_t}
\end{align}
where $i$ is the index of the
selected grid cell,
$Q_{i} \in \real^2$ is the center of that cell,
and $\tilde{O}_{ti} \in \real^2$ is the predicted offset for that cell at time $t$.
For single-future prediction, we use greedy search, namely $i=\argmax C_t$ over the belief state. For multi-future prediction, we use beam search in Section~\ref{sec:inference}.

This idea of combining classification
and regression is partially inspired by
object detection methods
(e.g., \cite{ren2015faster}).
\eat{
Different from prior works~\cite{bansal2018chauffeurnet,li2019way,manh2018scene,xue2018ss,liang2019peeking}, our model introduces an additional regression network for sequential fine-grained prediction.
}
It is worth noting that in concurrent work,
\cite{chai2019multipath} also designed a two-stage
model for trajectory forecasting.
However, their classification targets are pre-defined anchor trajectories.
Ours is not limited by the predefined anchors.

\vspace{-1mm}
\subsection{Training}
\vspace{-1mm}
Our model trains on the observed trajectory from time 1 to $h$ and predicts the future trajectories (in  $xy$-coordinates) from time $h+1$ to $T$.
We supervise this training by providing ground
truth targets for both the heatmap (belief state), $C_t^*$,
and regression offset map, $O_t^*$.
In particular, for the coarse decoder,
 the cross-entropy loss is used:
\begin{equation}
    \mathcal{L}_{cls} = -\frac{1}{T} \sum_{t=h+1}^{T} 
    \sum_{i \in G} C_{ti}^* \log(C_{ti})
\end{equation}
For the fine decoder, we use the smoothed $L_1$ loss
used in object detection~\cite{ren2015faster}:
\begin{equation}
 \mathcal{L}_{reg} = \frac{1}{T} \sum_{t=h+1}^{T} 
 \sum_{i \in G}
 \text{smooth}_{L_1}(O_{ti}^*, O_{ti})
\end{equation}
where 
$O_{ti}^* = L_t^* - Q_{i}$ 
is the delta between the true location and the center of the grid cell at $i$ and $L_t^*$ is the ground truth for $L_t$ in Eq.\eqref{eq:L_t}. We impose this loss on every cell to improve the robustness.

The final loss is then calculated using
\begin{align}
\mathcal{L}(\theta) = \mathcal{L}_{cls} + \lambda_1 \mathcal{L}_{reg} + \lambda_2 \|\theta\|_2^2
\end{align}
where $\lambda_2$ controls the $\ell_2$ regularization (weight decay),
and $\lambda_1=0.1$ is used
to balance the regression and classification losses.

Note that during training, when updating the 
RNN, we feed in the predicted
soft distribution
over locations, $C_{t}$. See Eq.~\eqref{eqn:fC}.
An alternative would be to feed in the true
values, $C_t^*$, \ie, use teacher forcing.
However, this is known to suffer
from problems \cite{Ranzato2016}.

\vspace{-1mm}
\subsection{Inference}\label{sec:inference}
\vspace{-1mm}
To generate multiple qualitatively distinct
 trajectories, we use the diverse beam search strategy from \cite{li2016simple}.
 To define this precisely,
 let $B_{t-1}$ be the beam at time $t-1$;
 this set contains $K$ trajectories (history
 selections) $M^k_{t-1}=\{\hat{Y}^k_1, \ldots , \hat{Y}^k_{t-1}\}$, $k \in[1 , K]$, where $\hat{Y}^k_t$ is an index in $G$,
 along with
 their accumulated log probabilities,
 $P^k_{t-1}$.
 Let $C^k_t=f(M^k_{t-1}) \in \mathbb{R}^{HW}$ be the coarse location output probability from Eq.~\eqref{eqn:fC} and \eqref{eqn:fH} at time $t$ given inputs $M^k_{t-1}$.
 
 The new beam is computed using

\begin{align}
 \small
    B_{t} = \text{topK}
    \left( \{
    P^k_{t-1} \!+\! \log(C^k_{t}(i)) \!+\! \gamma(i) | \forall i \in G, k \in [1 ,K]\}
    \right)
\end{align}
where $\gamma(i)$ is a diversity penalty term,
and we take the top $K$ elements from the set
produced by considering 
values with $k=1:K$.
If $K=1$, this reduces to greedy search.

 \eat{
 Here we use a single index $i \in [1, H \times W]$ to represent grid cells.
 Let $C_{t}(i)$ represents the probability of selecting cell $i$.
Define
\begin{align}
    P^K_{t} = {topK}^{K, H \times W}_{k=1, i=1} P^k_{t-1} + log(C^k_{t}(i)) + \gamma(i)
\end{align}
as the probability of the new beams at time $t$,
where $\gamma(i)$ is a diversity penalty term.
We select the grid cells at time $t$ based on the probabilities and get $M^k_{t}$ for the new beam selections.
If $K=1$, this reduces to greedy search.
}

Once we have computed the top $K$ future predictions,
we add the corresponding offset vectors to get
$K$ trajectories by $L^k_{t} \in \real^2$.
This constitutes the final output of our model.

\begin{figure*}[!t]
	\centering
	    \vspace{-4mm}
		\includegraphics[width=1.0\textwidth]{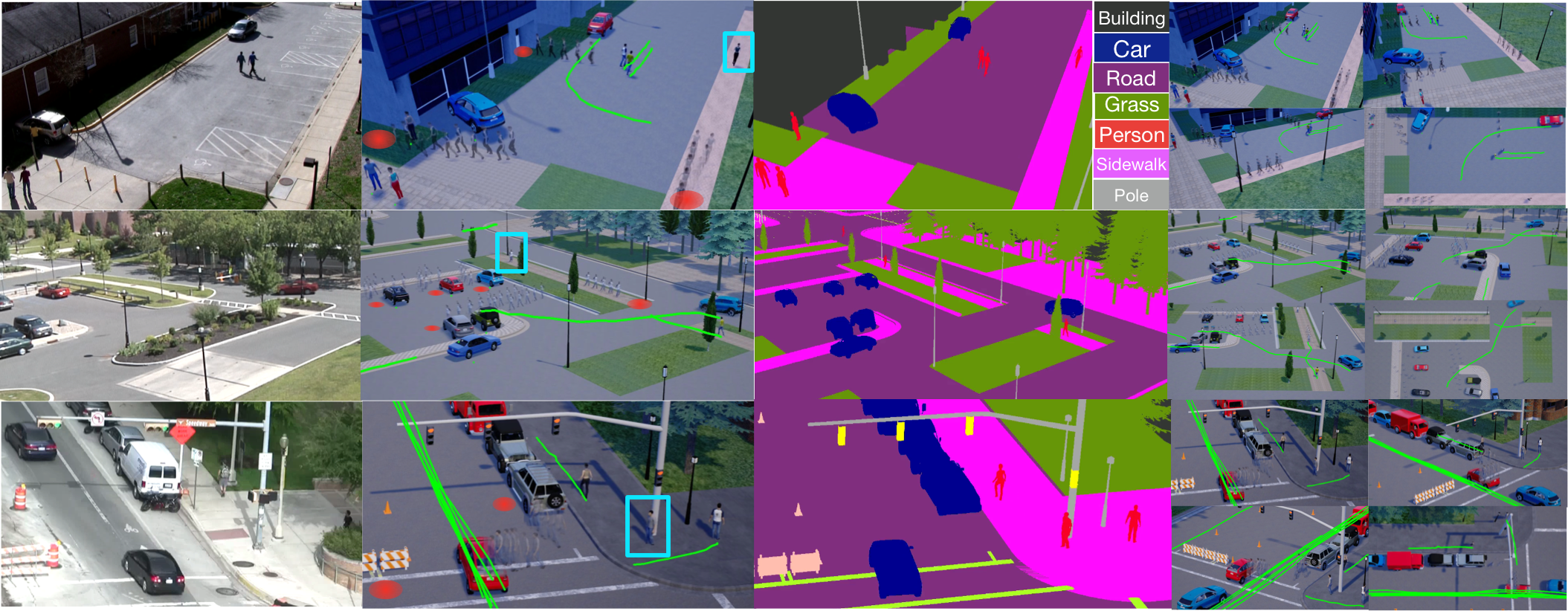}
		\vspace{-7mm}
	\caption{Visualization of the Forking Paths dataset. On the left is examples of the real videos and the second column shows the reconstructed scenes. The person in the blue bounding box is the controlled agent and multiple future trajectories annotated by humans are shown by overlaid person frames. 
	The red circles are the defined destinations. The green trajectories are future trajectories of the reconstructed uncontrolled agents. The scene semantic segmentation ground truth is shown in the third column and the last column shows all four camera views including the top-down view.} 
	\label{fig:dataset}
	\vspace{-6mm}
\end{figure*}

\vspace{-1mm}
\section{The Forking Paths Dataset}
\label{sec:dataset}
\vspace{-1mm}

In this section, we describe our human-annotated simulation dataset, called Forking Paths, for multi-future trajectory evaluation.

\noindent\textbf{Existing datasets.}
There are several real-world datasets for trajectory evaluation,
such as SDD~\cite{robicquet2016learning}, ETH/UCY~\cite{pellegrini2010improving,lerner2007crowds}, KITTI~\cite{geiger2013vision}, nuScenes~\cite{caesar2019nuscenes} and VIRAT/ActEV~\cite{2018trecvidawad,oh2011large}.
However, they all share the fundamental problem
that
one can only observe one out of many possible future trajectories sampled from the underlying distribution. This is broadly acknowledged in prior works~\cite{makansi2019overcoming,thiede2019analyzing,chai2019multipath,gupta2018social,rhinehart2019precog,rhinehart2018r2p2} but has not yet been addressed. 

The closest work to ours is the simulation used in~\cite{makansi2019overcoming, thiede2019analyzing,chai2019multipath}.
However,
these only contain artificial trajectories,
not human generated ones.
Also, they use a highly simplified  2D space, with pedestrians oversimplified as points and vehicles as blocks; no other scene semantics are provided.

\noindent\textbf{Reconstructing reality in simulator.}
In this work, we use CARLA~\cite{dosovitskiy2017carla}, a near-realistic open source simulator built on top of the Unreal Engine 4. 
Following prior simulation datasets~\cite{gaidon2016virtual,ros2016synthia}, we \textit{semi-automatically} reconstruct static scenes and their dynamic elements from the real-world videos in ETH/UCY and VIRAT/ActEV.
There are 4 scenes in ETH/UCY and 5 in VIRAT/ActEV. We exclude 2 cluttered scenes (UNIV \& 0002)
that we are not able to reconstruct in CARLA, leaving 7 static scenes in our dataset.

For dynamic movement of vehicle and pedestrian, we first convert the ground truth trajectory annotations from the real-world videos to the ground plane using the provided homography matrices.
We then match the real-world trajectories' origin to correct locations in the re-created scenes. 

\noindent\textbf{Human generation of plausible futures.}
We manually select sequences with more than one pedestrian.
We also require that
at least one pedestrian could have multiple plausible alternative destinations.
We insert plausible pedestrians into the scene to increase the diversity of the scenarios.
We then select one of the pedestrians to be the ``controlled agent'' (CA) for each sequence,  and set meaningful destinations within reach, like a car or an entrance of a building.
On average, each agent has about 3 destinations to move towards.
In total, 
we have 127 CAs from 7 scenes. We call each CA and their corresponding scene a scenario.

For each scenario,  
there are on average 5.9 human annotators to control the agent to the defined destinations. 
Specifically, they are asked to watch the first 5 seconds of video, from a first-person view (with the camera slightly behind the pedestrian) and/or an overhead view (to give more context). They are then asked to control the motion of the agent so that it moves towards the specified destination in a ``natural'' way, \eg, without colliding with other moving objects (whose motion is derived from the real videos, and is therefore unaware of the controlled agent).
The annotation is considered successful if the agent reached the destination without colliding within the time limit of 10.4 seconds. 
All final trajectories in our dataset are examined by humans to ensure reliability.

Note that our videos
are up to 15.2 seconds long.
This is slightly longer than previous works
(\eg \cite{alahi2016social,gupta2018social,liang2019peeking,sadeghian2018sophie,li2019way,zhang2019sr,zhao2019multi})
that use
3.2 seconds of observation and 
4.8 seconds for prediction.
(We use 10.4 seconds for the future
 to allow us to evaluate longer term forecasts.)

\noindent\textbf{Generating the data.}
Once we have collected human-generated trajectories, 750 in total after data cleaning, 
we render each one in four camera views (three 45-degree and one top-down view). Each camera view has 127 scenarios in total and each scenario has on average 5.9 future trajectories.
With CARLA,
we can also simulate different weather conditions,
although we did not do so in this work.
In addition to agent location, we collect ground truth for pixel-precise scene semantic segmentation from 13 classes including sidewalk, road, vehicle, pedestrian, \etc. See Fig.~\ref{fig:dataset}.

\vspace{-1mm}
\section{Experimental results}
\label{sec:exp}
\label{sec:results}
\vspace{-1mm}
This section evaluates various methods,
including our \emph{\fancyname} model,
for multi-future trajectory prediction
on the proposed Forking Paths dataset. 
To allow comparison with previous works,
we also evaluate our model on the challenging VIRAT/ActEV~\cite{2018trecvidawad,oh2011large} benchmark for single-future path prediction.

\subsection{Evaluation Metrics} 
\label{sec:metrics}

\noindent\textbf{Single-Future Evaluation.}
In real-world videos, each trajectory only has one sample of the future,
so models are evaluated on how well they predict that single trajectory.
Following prior work~\cite{liang2019peeking,alahi2016social,gupta2018social,sadeghian2018sophie,lee2017desire,hong2019rules,chai2019multipath,rhinehart2019precog}, we introduce two standard metrics for this setting.

Let ${Y}^i={Y}^i_{t=(h+1)\cdots T}$ be the ground truth trajectory of the $i$-th sample, and $\hat{Y}^i$ be the corresponding prediction.
We then employ two distance-based error metrics: 
\noindent i) \textit{Average Displacement Error} (ADE): the average Euclidean distance between the ground truth coordinates and the prediction coordinates over all time instants:
\begin{equation}
    \text{ADE} = \frac{ \sum^{N}_{i=1} \sum^{T}_{t=h+1} \lVert Y^i_t - \hat{Y}^i_t \rVert_{2}}{N \times (T-h)}
\end{equation}
\noindent ii) \textit{Final Displacement Error} (FDE): the Euclidean distance between the predicted points and the ground truth point at the final prediction time:
\begin{equation}
    \text{FDE} = \frac{ \sum^{N}_{i=1} \lVert Y^{i}_T - \hat{Y}^i_{T} \rVert_{2}}{N}
\end{equation}

\noindent\textbf{Multi-Future Evaluation.}
Let $Y^{ij}=Y^{ij}_{t=(h+1)\cdots T}$ be the $j$-th true future trajectory for the $i$-th test sample, for $\forall j \in [1,J]$,
and let $\hat{Y}^{ik}$ be the $k$'th sample from the predicted
distribution over trajectories, for $k \in [1,K]$.
Since there is no agreed-upon evaluation metric for this setting,
we simply extend the above metrics, as follows:
\noindent i) \textit{Minimum Average Displacement Error Given K Predictions} (minADE\textsubscript{K}): similar to the metric described in ~\cite{chai2019multipath, rhinehart2018r2p2,rhinehart2019precog, gupta2018social}, for each true trajectory $j$ in test sample $i$,
we select
the closest overall prediction (from the $K$ model predictions),
and then measure its average error:
\begin{equation}
    \text{minADE}_K = \frac{  \sum^{N}_{i=1} \sum^{J}_{j=1} \min_{k=1}^K \sum^{T}_{t=h+1} \lVert Y^{ij}_t - \hat{Y}^{ik}_t \rVert_{2}}{N \times (T-h) \times J}
\end{equation}

\noindent ii) \textit{Minimum Final Displacement Error Given K Predictions} (minFDE\textsubscript{K}): similar to minADE\textsubscript{K}, but we only consider the predicted points and the ground truth point at the final prediction time instant:
\begin{equation}
    \text{minFDE}_K = \frac{  \sum^{N}_{i=1} \sum^{J}_{j=1} \min_{k=1}^K \lVert Y^{ij}_{T} - \hat{Y}^{ik}_{T} \rVert_{2}}{N \times J}
\end{equation}

\noindent iii) \textit{Negative Log-Likelihood} (NLL):
Similar to NLL metrics used in~\cite{makansi2019overcoming,chai2019multipath}, we measure the
fit of ground-truth samples to the predicted distribution.

\subsection{Multi-Future Prediction on Forking Paths}\label{sec:exp-multi}

\begin{table*}
\vspace{-4mm}
\centering
\begin{tabular}{l|c|c|c|c|c}
\hline
\multirow{2}{*}{Method} & \multirow{2}{*}{Input Types} &  \multicolumn{2}{c|}{minADE\textsubscript{20}} & \multicolumn{2}{c}{minFDE\textsubscript{20}}  \\ \cline{3-6} 
 & & 45-degree         & top-down        & 45-degree         & top-down                   \\ \hline \hline
Linear & Traj. &  213.2    & 197.6    & 403.2    & 372.9   \\ 
LSTM  & Traj.              &  201.0 {\small$\pm$2.2} & 183.7 {\small$\pm$2.1} & 381.5 {\small$\pm$3.2} & 355.0 {\small$\pm$3.6}  \\ 
Social-LSTM \cite{alahi2016social}& Traj.         &  197.5 {\small$\pm$2.5} & 180.4 {\small$\pm$1.0}  & 377.0 {\small$\pm$3.6} & 350.3 {\small$\pm$2.3}   \\ 
Social-GAN (PV) \cite{gupta2018social}& Traj.         &  191.2 {\small$\pm$5.4} & 176.5 {\small$\pm$5.2}  & 351.9 {\small$\pm$11.4} & 335.0 {\small$\pm$9.4}   \\ 
Social-GAN (V) \cite{gupta2018social}& Traj.         &  187.1 {\small$\pm$4.7} & 172.7 {\small$\pm$3.9}  & 342.1 {\small$\pm$10.2} & 326.7 {\small$\pm$7.7}   \\ 
Next \cite{liang2019peeking}& Traj.+Bbox+RGB+Seg.   &    186.6 {\small$\pm$2.7} & 166.9 {\small$\pm$2.2}  & 360.0 {\small$\pm$7.2} & 326.6 {\small$\pm$5.0}    \\ 
 Ours& Traj.+Seg. & \textbf{168.9} {\small$\pm$2.1} & \textbf{157.7} {\small$\pm$2.5} & \textbf{333.8} {\small$\pm$3.7}  & \textbf{316.5} {\small$\pm$3.4}    \\ \hline
\end{tabular}
\vspace{-3mm}
\caption{Comparison of different methods on the Forking Paths dataset. Lower numbers are better. 
The numbers for the column labeled ``45 degrees'' are averaged
over 3 different 45-degree views.
For the input types, ``Traj.'', ``RGB'', ``Seg.'' and ``Bbox.'' mean the inputs are $xy$ coordinates, raw frames, semantic segmentations and bounding boxes of all objects in the scene, respectively.
All models are trained on real VIRAT/ActEV videos
and tested on synthetic (CARLA-rendered) videos.
}
\label{tab:exp-multi}
\vspace{-4mm}
\end{table*}
\noindent\textbf{Dataset \& Setups.} 
The proposed Forking Paths dataset in Section~\ref{sec:dataset} is used for multi-future trajectory prediction evaluation. 
Following the setting in previous works~\cite{liang2019peeking, alahi2016social, gupta2018social,alahi2016social, gupta2018social,sadeghian2018sophie, makansi2019overcoming}, 
we downsample the videos to 2.5 fps and extract person trajectories using code released in~\cite{liang2019peeking}, and let the models observe 3.2 seconds (8 frames) of the controlled agent before outputting trajectory coordinates in the pixel space. 
Since the length of the ground truth future trajectories are different, each model needs to predict the maximum length at test time but we evaluate the predictions using the actual length of each true trajectory.

\noindent\textbf{Baseline methods.} 
We compare our method with  two simple baselines, and three recent methods with released source code,
including  a recent model for multi-future prediction and the state-of-the-art model for single-future prediction:
\textbf{\textit{Linear}} is a single layer model that predicts the next coordinates using a linear regressor based on the previous input point.
\textbf{\textit{LSTM}} is a simple LSTM~\cite{hochreiter1997long} encoder-decoder model with coordinates input only. 
\textbf{\textit{Social LSTM}}~\cite{alahi2016social}: We use the open source implementation from {\footnotesize (\url{https://github.com/agrimgupta92/sgan/})}.
\textbf{\textit{Next}}~\cite{liang2019peeking} is the state-of-the-art method for single-future trajectory prediction on the VIRAT/ActEV dataset. We train the Next model without the activity labels for fair comparison using the code from {\footnotesize (\url{https://github.com/google/next-prediction/})}.
\textbf{\textit{Social GAN}}~\cite{gupta2018social} is a recent multi-future trajectory prediction model trained using Minimum over N (MoN) loss. We train two model variants (called PV and V) detailed in the paper using the code from~\cite{gupta2018social} .

All models are trained on real videos
(from VIRAT/ActEV -- see Section~\ref{sec:exp-virat} for  details)
and tested on our synthetic videos
(with CARLA-generated pixels,
and annotator-generated trajectories).
Most models just use trajectory data as input,
except for our model
(which uses trajectory and semantic segmentation)
and Next
(which uses trajectory, bounding box,
semantic segmentation, and RGB frames).

\noindent\textbf{Implementation Details.}
We use ConvLSTM~\cite{xingjian2015convolutional} cell for both the encoder and decoder. 
The embedding size is set to 32, and the hidden sizes for the encoder and decoder are both 256. 
The scene semantic segmentation features are extracted from the deeplab model~\cite{chen2017deeplab}, pretrained on the ADE-20k~\cite{zhou2017scene} dataset.
We use Adadelta optimizer~\cite{zeiler2012adadelta} with an initial learning rate of 0.3 and weight decay of 0.001. 
Other hyper-parameters for the baselines are the same to the ones in ~\cite{gupta2018social, liang2019peeking}.
We evaluate the top $K=20$ predictions for multi-future trajectories. For the models that only output a single trajectory, including Linear, LSTM, Social-LSTM, and Next, we duplicate the output for $K$ times before evaluating. For Social-GAN, we use $K$ different random noise inputs to get the predictions. For our model, we use diversity beam search~\cite{li2016simple,plotz2018neural} as described in Section~\ref{sec:inference}.

\begin{table}[]
\centering
\begin{tabular}{l||c|c|c}
\hline
Method & $T_{pred}=1$   & $T_{pred}=2$ & $T_{pred}=3$  \\ \hline \hline
(PV) [14]    &  10.08 {\scriptsize$\pm$0.25} & 17.28 {\scriptsize$\pm$0.42} &23.34 {\scriptsize$\pm$0.47}  \\ 
(V) [14]    &  9.95 {\scriptsize$\pm$0.35} & 17.38 {\scriptsize$\pm$0.49} &23.24 {\scriptsize$\pm$0.54} \\ 
Next [27]  & 8.32 {\scriptsize$\pm$0.10} & 14.98 {\scriptsize$\pm$0.19} &22.71 {\scriptsize$\pm$0.11} \\ 
 Ours & \textbf{2.22} {\scriptsize$\pm$0.54} & \textbf{4.46} {\scriptsize$\pm$1.33} & \textbf{8.14} {\scriptsize$\pm$2.81} \\ \hline
\end{tabular}
\vspace{-3mm}
\caption{Negative Log-likelihood comparison of different methods on the Forking Paths dataset. For methods that output multiple trajectories, we quantize the xy-coordinates into the same grid as our method and get a normalized probability distribution prediction.} 
\vspace{-6mm}
\label{tab:exp-nll}
\end{table}

\noindent\textbf{Quantitative Results.} 
Table~\ref{tab:exp-multi} lists the multi-future evaluation results, where we divide the  evaluation according to the viewing angle of camera, 45-degree vs. top-down view. 
We repeat all experiments (except ``linear'') 5 times with random initialization to produce the mean and standard deviation values.
As we see, our model outperforms baselines in all metrics and it performs significantly better on the minADE metric, which suggests better prediction quality over all time instants.
Notably, our model outperforms Social GAN by a large margin of at least 8 points on all metrics.
We also measure the standard negative log-likelihood (NLL) metric for the top methods in Table~\ref{tab:exp-nll}.

\begin{figure*}[!t]
\vspace{-4mm}
	\centering
		\includegraphics[width=1.0\textwidth]{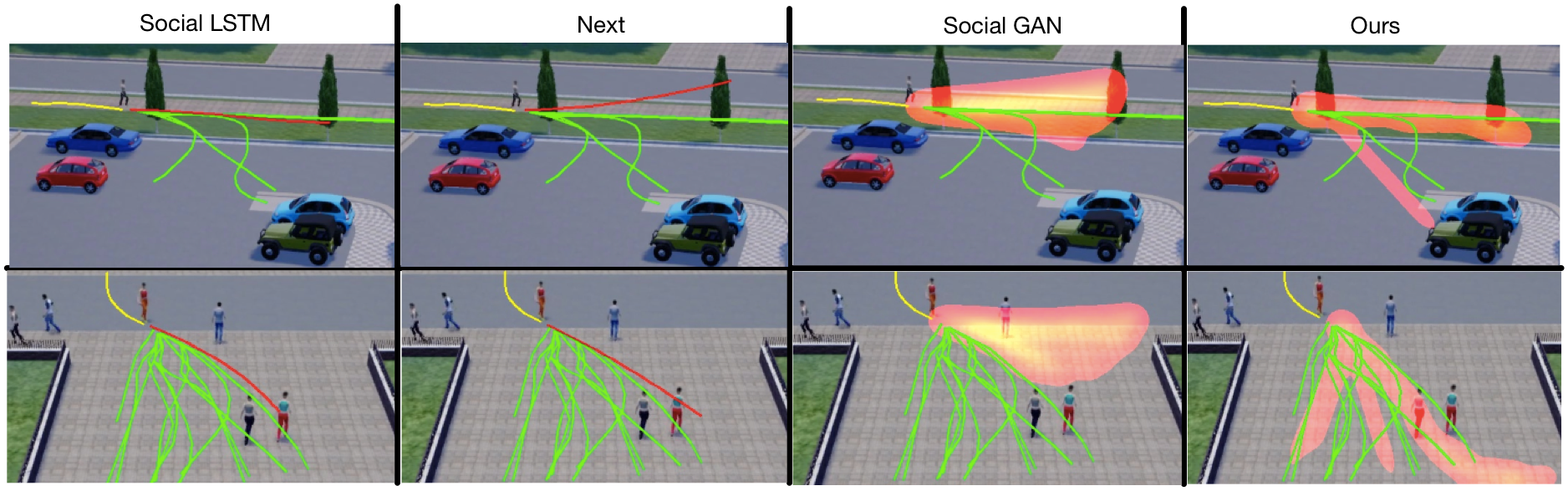}
		\vspace{-7mm}
	\caption{Qualitative analysis. The red trajectories are single-future method predictions and the yellow-orange heatmaps are multi-future method predictions. The yellow trajectories are observations and the green ones are ground truth multi-future trajectories. See text for details.}  
	\label{fig:qual}
	\vspace{-6mm}
\end{figure*}

\noindent\textbf{Qualitative analysis.} 
We visualize some outputs of the top 4 methods in Fig.~\ref{fig:qual}.
In each image, the yellow trajectories are the history trajectory of each controlled agent (derived from real 
video data)
and the green trajectories are the ground truth
 future trajectories from human annotators.
The predicted trajectories are shown in yellow-orange heatmaps for multi-future prediction methods, and in red lines for single-future prediction methods. 
As we see, our model correctly generally
puts probability
mass where there is data, and does not ``waste''
probability mass where there is no data.

\begin{figure}[!t]
	\centering
	    \vspace{-2mm}
		\includegraphics[width=0.47\textwidth]{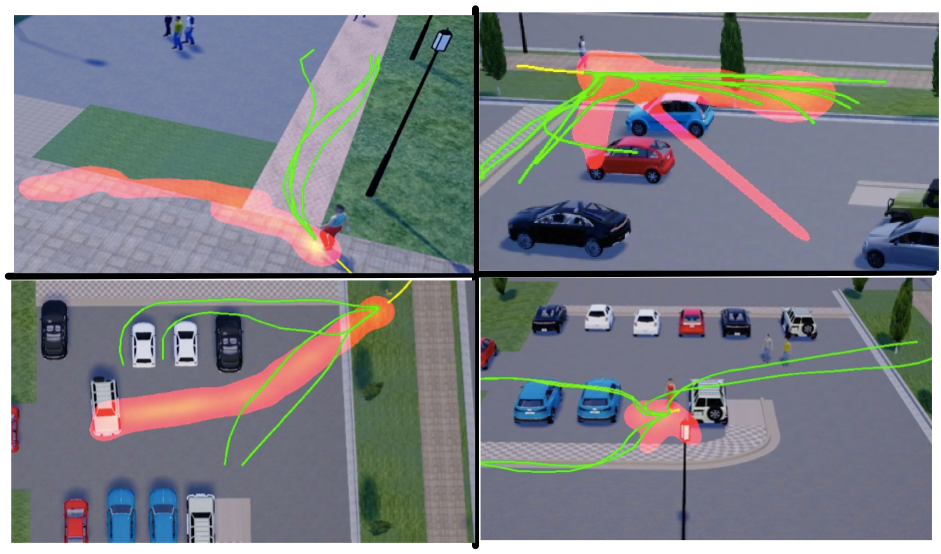}
		\vspace{-4mm}
	\caption{Error analysis. See text for details.}  
	\label{fig:error_analysis}
	\vspace{-4mm}
\end{figure}

\noindent\textbf{Error analysis.}
We show some typical errors our model makes in Fig.~\ref{fig:error_analysis}.
The first image shows our model misses the correct direction,
perhaps due to lack of diversity in our sampling procedure.
The second image shows our model 
sometimes predicts the person will ``go through'' the car
(diagonal red beam)
instead of going around it. This may be addressed by adding more training examples of ``going around'' obstacles.
The third image shows
our model predicts the person will go to a moving car.
This is due to the lack of modeling of the dynamics of other far-away agents in the scene. 
The fourth image shows a hard case where the person just exits the vehicle and there is no indication of where they will go next (so our model ``backs off'' to a sensible
``stay nearby'' prediction).
We leave solutions to these problems to future work.

\subsection{Single-Future Prediction on VIRAT/ActEV}
\label{sec:exp-virat}

\noindent\textbf{Dataset \& Setups.} NIST released VIRAT/ActEV~\cite{2018trecvidawad} for activity detection research in streaming videos in 2018. 
This dataset is a new version of the VIRAT~\cite{oh2011large} dataset, with more videos and annotations. The length of videos with publicly available annotations is about 4.5 hours.
Following~\cite{liang2019peeking}, we use the official training set for training and the official validation set for testing. Other setups are the same as in Section~\ref{sec:exp-multi},
except 
we use the single-future evaluation metric.

\noindent\textbf{Quantitative Results.} 
Table~\ref{tab:exp-single} (first column)
shows the evaluation results. 
As we see, our model achieves state-of-the-art performance.
The improvement is especially large on Final Displacement Error (FDE) metric, attributing to the coarse location decoder that helps regulate the model prediction for long-term prediction.
The gain shows that our model does well at both
single future prediction (on real data)
and multiple future prediction on our quasi-synthetic data.

\noindent\textbf{Generalizing from simulation to real-world.} As described in Section~\ref{sec:dataset}, we generate simulation data first by reconstructing from real-world videos.
To verify the quality of the reconstructed data,
and the efficacy of learning from simulation videos, we train 
all the models on the simulation videos 
derived from the real data.
We then evaluate on the real test set of VIRAT/ActEV.
As we see from the right column in Table~\ref{tab:exp-single},
all models do worse in this scenario,
due to the difference between synthetic and real data.
We find  the  performance  ranking  of  different  methods  are consistent between the real and our simulation training data.  This suggests the errors mainly coming from the model,  and  substantiates  the  rationality  of  using  the  proposed dataset to compare the relative performance of different methods.

There are two sources of error.
The synthetic trajectory data only contains about 60\%
of the real trajectory data, due to difficulties
reconstructing all the real data in the simulator.
In addition, the synthetic images are
not photo realistic.
Thus
methods (such as Next \cite{liang2019peeking})
that rely on RGB input obviously suffer the most,
since they have never been trained on ``real pixels''.
Our method, which uses trajectories plus high level
semantic segmentations (which transfers from synthetic
to real more easily)
suffers the least drop in performance,
showing its robustness to ``domain shift''. See Table~\ref{tab:exp-multi} for input source comparison between methods.

\begin{table}[]
\centering
\small
\begin{tabular}{l||c||c}
\hline
Method & Trained on Real.   & Trained on Sim.        \\ \hline \hline
Linear &  32.19 / 60.92 & 48.65 / 90.84 \\ 
LSTM                &  23.98 / 44.97 & 28.45 / 53.01 \\ 
Social-LSTM \cite{alahi2016social}         &  23.10 / 44.27 & 26.72 / 51.26 \\ 
Social-GAN (V) \cite{gupta2018social}       &  30.40 / 61.93 & 36.74 / 73.22 \\ 
Social-GAN (PV) \cite{gupta2018social}         &  30.42 / 60.70 & 36.48 / 72.72 \\ 
Next ~\cite{liang2019peeking}   & 19.78 / 42.43 & 27.38 / 62.11 \\ 
 Ours & \textbf{18.51} / \textbf{35.84} & \textbf{22.94} / \textbf{43.35} \\ \hline
\end{tabular}
\vspace{-3mm}
\caption{Comparison of different methods on the VIRAT/ActEV dataset.
We report ADE/FDE metrics.
First column  is for models trained on real video training set and
second column 
 is for models trained on the simulated version of this
 dataset.} 
\label{tab:exp-single}
\vspace{-6mm}
\end{table}

\vspace{-1mm}
\subsection{Ablation Experiments}
\label{sec:ablation}
\vspace{-1mm}
We test various ablations of our model
on both the single-future and multi-future trajectory prediction to substantiate our design decisions.
Results are shown in Table~\ref{tab:exp-ablation}, where the ADE/FDE metrics are shown in the ``single-future'' column and minADE\textsubscript{20}/minFDE\textsubscript{20} metrics (averaged across all views) in the ``multi-future'' column. We verify three of our key designs by leaving the module out from the full model.

(1) \textit{Spatial Graph:} Our model is built on top of a spatial 2D graph that uses graph attention to model the scene features. We train model without the spatial graph.
As we see, the performance drops on both tasks.
(2) \textit{Fine location decoder:} We test our model without the fine location decoder and only use the grid center as the coordinate output.
As we see, the significant performance drops on both tasks verify the efficacy of this new module proposed in our study. 
(3) \textit{Multi-scale grid:} As mentioned in Section~\ref{sec:approach}, we utilize two different grid scales (36 $\times$ 18) and (18 $\times$ 9) in training.
We see that performance is slightly worse if we only use the fine scale (36 $\times$ 18) .

\begin{table}
\centering
\small
\begin{tabular}{l||c|c}
\hline
Method                             & Single-Future & Multi-Future \\ \hline
Our full model      &   18.51 / 35.84  &    166.1 / 329.5    \\ \hline
No spatial graph         &   28.68 / 49.87  & 184.5 / 363.2     \\ 
No fine location decoder                 &   53.62 / 83.57   &  232.1 / 468.6 \\ 
No multi-scale grid                       &  21.09 / 38.45     &    171.0 / 344.4     \\ \hline
\end{tabular}
\vspace{-3mm}
\caption{Performance on ablated versions of our model on single and multi-future trajectory prediction. 
Lower numbers are better.}
\label{tab:exp-ablation}
\vspace{-6mm}
\end{table}

\vspace{-2mm}
\section{Conclusion}
\label{sec:concl}
\vspace{-2mm}
In this paper,
we have introduced the Forking Paths dataset, 
and the \textit{\fancyname} model for multi-future forecasting.
Our study is the first to provide a quantitative benchmark and evaluation methodology for 
multi-future trajectory prediction by using human annotators to create a variety
of trajectory continuations under the identical past.
Our model utilizes multi-scale location decoders with 
graph attention model to predict multiple future locations.
We have shown that our method achieves state-of-the-art performance on two
challenging benchmarks: the large-scale real video dataset and our proposed multi-future trajectory dataset.
We believe our dataset, together with our models, will facilitate future research and applications on multi-future prediction.

\noindent\textbf{Acknowledgements} 
This research was supported by NSF grant IIS-1650994, the financial assistance award 60NANB17D156 from NIST and a Baidu Scholarship.
This work was also supported by IARPA
via DOI/IBC contract number D17PC00340.
The views and
conclusions contained herein are those of the authors and
should not be interpreted as necessarily representing the official policies or endorsements, either expressed or implied,
of IARPA, NIST, DOI/IBC, the National Science Foundation, Baidu, or the U.S. Government.

{
\bibliographystyle{ieee_fullname}
\bibliography{egbib}

\begin{thebibliography}{10}\itemsep=-1pt

\bibitem{alahi2016social}
Alexandre Alahi, Kratarth Goel, Vignesh Ramanathan, Alexandre Robicquet, Li
  Fei-Fei, and Silvio Savarese.
\newblock Social lstm: Human trajectory prediction in crowded spaces.
\newblock In {\em CVPR}, 2016.

\bibitem{amirian2019social}
Javad Amirian, Jean-Bernard Hayet, and Julien Pettr{\'e}.
\newblock Social ways: Learning multi-modal distributions of pedestrian
  trajectories with gans.
\newblock In {\em CVPRW}, 2019.

\bibitem{2018trecvidawad}
George Awad, Asad Butt, Keith Curtis, Jonathan Fiscus, Afzal Godil, Alan~F.
  Smeaton, Yvette Graham, Wessel Kraaij, Georges Quénot, Joao Magalhaes, David
  Semedo, and Saverio Blasi.
\newblock Trecvid 2018: Benchmarking video activity detection, video captioning
  and matching, video storytelling linking and video search.
\newblock In {\em TRECVID}, 2018.

\bibitem{bansal2018chauffeurnet}
Mayank Bansal, Alex Krizhevsky, and Abhijit Ogale.
\newblock Chauffeurnet: Learning to drive by imitating the best and
  synthesizing the worst.
\newblock {\em arXiv preprint arXiv:1812.03079}, 2018.

\bibitem{caesar2019nuscenes}
Holger Caesar, Varun Bankiti, Alex~H Lang, Sourabh Vora, Venice~Erin Liong,
  Qiang Xu, Anush Krishnan, Yu Pan, Giancarlo Baldan, and Oscar Beijbom.
\newblock nuscenes: A multimodal dataset for autonomous driving.
\newblock {\em arXiv preprint arXiv:1903.11027}, 2019.

\bibitem{chai2019multipath}
Yuning Chai, Benjamin Sapp, Mayank Bansal, and Dragomir Anguelov.
\newblock Multipath: Multiple probabilistic anchor trajectory hypotheses for
  behavior prediction.
\newblock {\em arXiv preprint arXiv:1910.05449}, 2019.

\bibitem{chang2019argoverse}
Ming-Fang Chang, John Lambert, Patsorn Sangkloy, Jagjeet Singh, Slawomir Bak,
  Andrew Hartnett, De Wang, Peter Carr, Simon Lucey, Deva Ramanan, et~al.
\newblock Argoverse: 3d tracking and forecasting with rich maps.
\newblock In {\em CVPR}, 2019.

\bibitem{chen2017deeplab}
Liang-Chieh Chen, George Papandreou, Iasonas Kokkinos, Kevin Murphy, and Alan~L
  Yuille.
\newblock Deeplab: Semantic image segmentation with deep convolutional nets,
  atrous convolution, and fully connected crfs.
\newblock {\em IEEE transactions on pattern analysis and machine intelligence},
  40(4):834--848, 2017.

\bibitem{das2018embodied}
Abhishek Das, Samyak Datta, Georgia Gkioxari, Stefan Lee, Devi Parikh, and
  Dhruv Batra.
\newblock Embodied question answering.
\newblock In {\em CVPRW}, 2018.

\bibitem{de2017procedural}
C{\'e}sar~Roberto de Souza, Adrien Gaidon, Yohann Cabon, and Antonio~Manuel
  L{\'o}pez.
\newblock Procedural generation of videos to train deep action recognition
  networks.
\newblock In {\em CVPR}, 2017.

\bibitem{dosovitskiy2017carla}
Alexey Dosovitskiy, German Ros, Felipe Codevilla, Antonio Lopez, and Vladlen
  Koltun.
\newblock Carla: An open urban driving simulator.
\newblock {\em arXiv preprint arXiv:1711.03938}, 2017.

\bibitem{gaidon2016virtual}
Adrien Gaidon, Qiao Wang, Yohann Cabon, and Eleonora Vig.
\newblock Virtual worlds as proxy for multi-object tracking analysis.
\newblock In {\em CVPR}, 2016.

\bibitem{geiger2013vision}
Andreas Geiger, Philip Lenz, Christoph Stiller, and Raquel Urtasun.
\newblock Vision meets robotics: The kitti dataset.
\newblock {\em The International Journal of Robotics Research},
  32(11):1231--1237, 2013.

\bibitem{goodfellow2014generative}
Ian Goodfellow, Jean Pouget-Abadie, Mehdi Mirza, Bing Xu, David Warde-Farley,
  Sherjil Ozair, Aaron Courville, and Yoshua Bengio.
\newblock Generative adversarial nets.
\newblock In {\em NeurIPS}, 2014.

\bibitem{gupta2018social}
Agrim Gupta, Justin Johnson, Silvio Savarese, Li Fei-Fei, and Alexandre Alahi.
\newblock Social gan: Socially acceptable trajectories with generative
  adversarial networks.
\newblock In {\em CVPR}, 2018.

\bibitem{heess2017emergence}
Nicolas Heess, Srinivasan Sriram, Jay Lemmon, Josh Merel, Greg Wayne, Yuval
  Tassa, Tom Erez, Ziyu Wang, SM Eslami, Martin Riedmiller, et~al.
\newblock Emergence of locomotion behaviours in rich environments.
\newblock {\em arXiv preprint arXiv:1707.02286}, 2017.

\bibitem{hochreiter1997long}
Sepp Hochreiter and J{\"u}rgen Schmidhuber.
\newblock Long short-term memory.
\newblock {\em Neural computation}, 9(8):1735--1780, 1997.

\bibitem{hong2019rules}
Joey Hong, Benjamin Sapp, and James Philbin.
\newblock Rules of the road: Predicting driving behavior with a convolutional
  model of semantic interactions.
\newblock In {\em CVPR}, 2019.

\bibitem{kalman1960new}
RE Kalman.
\newblock A new approach to linear filtering and prediction problems.
\newblock {\em Trans. ASME, D}, 82:35--44, 1960.

\bibitem{kitani2012activity}
Kris~M Kitani, Brian~D Ziebart, James~Andrew Bagnell, and Martial Hebert.
\newblock Activity forecasting.
\newblock In {\em ECCV}, 2012.

\bibitem{kooij2014context}
Julian Francisco~Pieter Kooij, Nicolas Schneider, Fabian Flohr, and Dariu~M
  Gavrila.
\newblock Context-based pedestrian path prediction.
\newblock In {\em ECCV}, 2014.

\bibitem{lazebnik2006beyond}
Svetlana Lazebnik, Cordelia Schmid, and Jean Ponce.
\newblock Beyond bags of features: Spatial pyramid matching for recognizing
  natural scene categories.
\newblock In {\em CVPR}, 2006.

\bibitem{lee2017desire}
Namhoon Lee, Wongun Choi, Paul Vernaza, Christopher~B Choy, Philip~HS Torr, and
  Manmohan Chandraker.
\newblock Desire: Distant future prediction in dynamic scenes with interacting
  agents.
\newblock In {\em CVPR}, 2017.

\bibitem{lerner2007crowds}
Alon Lerner, Yiorgos Chrysanthou, and Dani Lischinski.
\newblock Crowds by example.
\newblock In {\em Computer Graphics Forum}, pages 655--664. Wiley Online
  Library, 2007.

\bibitem{li2016simple}
Jiwei Li, Will Monroe, and Dan Jurafsky.
\newblock A simple, fast diverse decoding algorithm for neural generation.
\newblock {\em arXiv preprint arXiv:1611.08562}, 2016.

\bibitem{li2019way}
Yuke Li.
\newblock Which way are you going? imitative decision learning for path
  forecasting in dynamic scenes.
\newblock In {\em CVPR}, 2019.

\bibitem{liang2017event}
Junwei Liang, Desai Fan, Han Lu, Poyao Huang, Jia Chen, Lu Jiang, and Alexander
  Hauptmann.
\newblock An event reconstruction tool for conflict monitoring using social
  media.
\newblock In {\em AAAI}, 2017.

\bibitem{liang2019focal}
Junwei Liang, Lu Jiang, Liangliang Cao, Yannis Kalantidis, Li-Jia Li, and
  Alexander~G Hauptmann.
\newblock Focal visual-text attention for memex question answering.
\newblock {\em IEEE transactions on pattern analysis and machine intelligence},
  41(8):1893--1908, 2019.

\bibitem{liang2018focal}
Junwei Liang, Lu Jiang, Liangliang Cao, Li-Jia Li, and Alexander~G Hauptmann.
\newblock Focal visual-text attention for visual question answering.
\newblock In {\em CVPR}, 2018.

\bibitem{liang2019peeking}
Junwei Liang, Lu Jiang, Juan~Carlos Niebles, Alexander~G Hauptmann, and Li
  Fei-Fei.
\newblock Peeking into the future: Predicting future person activities and
  locations in videos.
\newblock In {\em CVPR}, 2019.

\bibitem{lin2017feature}
Tsung-Yi Lin, Piotr Doll{\'a}r, Ross Girshick, Kaiming He, Bharath Hariharan,
  and Serge Belongie.
\newblock Feature pyramid networks for object detection.
\newblock In {\em CVPR}, 2017.

\bibitem{luber2010people}
Matthias Luber, Johannes~A Stork, Gian~Diego Tipaldi, and Kai~O Arras.
\newblock People tracking with human motion predictions from social forces.
\newblock In {\em ICRA}, 2010.

\bibitem{ma2017forecasting}
Wei-Chiu Ma, De-An Huang, Namhoon Lee, and Kris~M Kitani.
\newblock Forecasting interactive dynamics of pedestrians with fictitious play.
\newblock In {\em CVPR}, 2017.

\bibitem{makansi2019overcoming}
Osama Makansi, Eddy Ilg, Ozgun Cicek, and Thomas Brox.
\newblock Overcoming limitations of mixture density networks: A sampling and
  fitting framework for multimodal future prediction.
\newblock In {\em CVPR}, 2019.

\bibitem{manh2018scene}
Huynh Manh and Gita Alaghband.
\newblock Scene-lstm: A model for human trajectory prediction.
\newblock {\em arXiv preprint arXiv:1808.04018}, 2018.

\bibitem{oh2011large}
Sangmin Oh, Anthony Hoogs, Amitha Perera, Naresh Cuntoor, Chia-Chih Chen,
  Jong~Taek Lee, Saurajit Mukherjee, JK Aggarwal, Hyungtae Lee, Larry Davis,
  et~al.
\newblock A large-scale benchmark dataset for event recognition in surveillance
  video.
\newblock In {\em CVPR}, 2011.

\bibitem{pellegrini2010improving}
Stefano Pellegrini, Andreas Ess, and Luc Van~Gool.
\newblock Improving data association by joint modeling of pedestrian
  trajectories and groupings.
\newblock In {\em ECCV}, 2012.

\bibitem{plotz2018neural}
Tobias Pl{\"o}tz and Stefan Roth.
\newblock Neural nearest neighbors networks.
\newblock In {\em NeurIPS}, 2018.

\bibitem{qiu2017unrealcv}
Weichao Qiu, Fangwei Zhong, Yi Zhang, Siyuan Qiao, Zihao Xiao, Tae~Soo Kim, and
  Yizhou Wang.
\newblock Unrealcv: Virtual worlds for computer vision.
\newblock In {\em ACM Multimedia}, 2017.

\bibitem{Ranzato2016}
Marc'Aurelio Ranzato, Sumit Chopra, Michael Auli, and Wojciech Zaremba.
\newblock Sequence level training with recurrent neural networks.
\newblock {\em arXiv preprint arXiv:1511.06732}, 2015.

\bibitem{ren2015faster}
Shaoqing Ren, Kaiming He, Ross Girshick, and Jian Sun.
\newblock Faster r-cnn: Towards real-time object detection with region proposal
  networks.
\newblock In {\em NeurIPS}, 2015.

\bibitem{rhinehart2017first}
Nicholas Rhinehart and Kris~M Kitani.
\newblock First-person activity forecasting with online inverse reinforcement
  learning.
\newblock In {\em ICCV}, 2017.

\bibitem{rhinehart2018r2p2}
Nicholas Rhinehart, Kris~M Kitani, and Paul Vernaza.
\newblock R2p2: A reparameterized pushforward policy for diverse, precise
  generative path forecasting.
\newblock In {\em ECCV}, 2018.

\bibitem{rhinehart2019precog}
Nicholas Rhinehart, Rowan McAllister, Kris Kitani, and Sergey Levine.
\newblock Precog: Prediction conditioned on goals in visual multi-agent
  settings.
\newblock {\em arXiv preprint arXiv:1905.01296}, 2019.

\bibitem{richter2016playing}
Stephan~R Richter, Vibhav Vineet, Stefan Roth, and Vladlen Koltun.
\newblock Playing for data: Ground truth from computer games.
\newblock In {\em ECCV}, 2016.

\bibitem{robicquet2016learning}
Alexandre Robicquet, Amir Sadeghian, Alexandre Alahi, and Silvio Savarese.
\newblock Learning social etiquette: Human trajectory understanding in crowded
  scenes.
\newblock In {\em ECCV}, 2016.

\bibitem{ros2016synthia}
German Ros, Laura Sellart, Joanna Materzynska, David Vazquez, and Antonio~M
  Lopez.
\newblock The synthia dataset: A large collection of synthetic images for
  semantic segmentation of urban scenes.
\newblock In {\em CVPR}, 2016.

\bibitem{sadeghian2017tracking}
Amir Sadeghian, Alexandre Alahi, and Silvio Savarese.
\newblock Tracking the untrackable: Learning to track multiple cues with
  long-term dependencies.
\newblock In {\em ICCV}, 2017.

\bibitem{sadeghian2018sophie}
Amir Sadeghian, Vineet Kosaraju, Ali Sadeghian, Noriaki Hirose, and Silvio
  Savarese.
\newblock Sophie: An attentive gan for predicting paths compliant to social and
  physical constraints.
\newblock {\em arXiv preprint arXiv:1806.01482}, 2018.

\bibitem{sadeghian2018car}
Amir Sadeghian, Ferdinand Legros, Maxime Voisin, Ricky Vesel, Alexandre Alahi,
  and Silvio Savarese.
\newblock Car-net: Clairvoyant attentive recurrent network.
\newblock In {\em ECCV}, 2018.

\bibitem{shah2018airsim}
Shital Shah, Debadeepta Dey, Chris Lovett, and Ashish Kapoor.
\newblock Airsim: High-fidelity visual and physical simulation for autonomous
  vehicles.
\newblock In {\em Field and service robotics}, pages 621--635. Springer, 2018.

\bibitem{sun2019stochastic}
Chen Sun, Per Karlsson, Jiajun Wu, Joshua~B Tenenbaum, and Kevin Murphy.
\newblock Stochastic prediction of multi-agent interactions from partial
  observations.
\newblock {\em arXiv preprint arXiv:1902.09641}, 2019.

\bibitem{tang2019multiple}
Yichuan~Charlie Tang and Ruslan Salakhutdinov.
\newblock Multiple futures prediction.
\newblock {\em arXiv preprint arXiv:1911.00997}, 2019.

\bibitem{thiede2019analyzing}
Luca~Anthony Thiede and Pratik~Prabhanjan Brahma.
\newblock Analyzing the variety loss in the context of probabilistic trajectory
  prediction.
\newblock {\em arXiv preprint arXiv:1907.10178}, 2019.

\bibitem{velivckovic2017graph}
Petar Veli{\v{c}}kovi{\'c}, Guillem Cucurull, Arantxa Casanova, Adriana Romero,
  Pietro Lio, and Yoshua Bengio.
\newblock Graph attention networks.
\newblock {\em arXiv preprint arXiv:1710.10903}, 2017.

\bibitem{wang2019eidetic}
Yunbo Wang, Lu Jiang, Ming-Hsuan Yang, Li-Jia Li, Mingsheng Long, and Li
  Fei-Fei.
\newblock Eidetic 3d lstm: A model for video prediction and beyond.
\newblock In {\em ICLR}, 2019.

\bibitem{wu2019revisiting}
Yu Wu, Lu Jiang, and Yi Yang.
\newblock Revisiting embodiedqa: A simple baseline and beyond.
\newblock {\em arXiv preprint arXiv:1904.04166}, 2019.

\bibitem{xingjian2015convolutional}
SHI Xingjian, Zhourong Chen, Hao Wang, Dit-Yan Yeung, Wai-Kin Wong, and
  Wang-chun Woo.
\newblock Convolutional lstm network: A machine learning approach for
  precipitation nowcasting.
\newblock In {\em NeurIPS}, 2015.

\bibitem{xue2018ss}
Hao Xue, Du~Q Huynh, and Mark Reynolds.
\newblock Ss-lstm: A hierarchical lstm model for pedestrian trajectory
  prediction.
\newblock In {\em WACV}, 2018.

\bibitem{yagi2018future}
Takuma Yagi, Karttikeya Mangalam, Ryo Yonetani, and Yoichi Sato.
\newblock Future person localization in first-person videos.
\newblock In {\em CVPR}, 2018.

\bibitem{zeiler2012adadelta}
Matthew~D Zeiler.
\newblock Adadelta: an adaptive learning rate method.
\newblock {\em arXiv preprint arXiv:1212.5701}, 2012.

\bibitem{zhang2019sr}
Pu Zhang, Wanli Ouyang, Pengfei Zhang, Jianru Xue, and Nanning Zheng.
\newblock Sr-lstm: State refinement for lstm towards pedestrian trajectory
  prediction.
\newblock In {\em CVPR}, 2019.

\bibitem{zhang2015fast}
Yiwei Zhang, Graham~M Gibson, Rebecca Hay, Richard~W Bowman, Miles~J Padgett,
  and Matthew~P Edgar.
\newblock A fast 3d reconstruction system with a low-cost camera accessory.
\newblock {\em Scientific reports}, 5:10909, 2015.

\bibitem{zhao2019multi}
Tianyang Zhao, Yifei Xu, Mathew Monfort, Wongun Choi, Chris Baker, Yibiao Zhao,
  Yizhou Wang, and Ying~Nian Wu.
\newblock Multi-agent tensor fusion for contextual trajectory prediction.
\newblock In {\em CVPR}, 2019.

\bibitem{zhou2017scene}
Bolei Zhou, Hang Zhao, Xavier Puig, Sanja Fidler, Adela Barriuso, and Antonio
  Torralba.
\newblock Scene parsing through ade20k dataset.
\newblock In {\em CVPR}, 2017.

\bibitem{zhu2017target}
Yuke Zhu, Roozbeh Mottaghi, Eric Kolve, Joseph~J Lim, Abhinav Gupta, Li
  Fei-Fei, and Ali Farhadi.
\newblock Target-driven visual navigation in indoor scenes using deep
  reinforcement learning.
\newblock In {\em ICRA}, 2017.

\end{thebibliography}
}
\end{document}